\documentclass{article}

\usepackage{docmute}

\usepackage{arxiv}

\usepackage[utf8]{inputenc}
\usepackage[T1]{fontenc}

\usepackage{amsmath}
\usepackage{amssymb}
\usepackage{amsfonts}
\usepackage{amsthm}

\usepackage{comment}
\usepackage{color}

\usepackage{graphicx}

\usepackage{physics}

\usepackage{cite}
\usepackage{hyperref}
\usepackage{url}

\usepackage[whole,substmingoth]{bxcjkjatype}

\usepackage{enumitem}

\usepackage{multicol}
\usepackage{multirow}

\usepackage{makecell}
\usepackage{tabularx}
\newcolumntype{Y}{>{\centering\arraybackslash}X}
\newcolumntype{P}{>{\raggedleft\arraybackslash}X}

\usepackage{svg}

\usepackage{subcaption}

\expandafter\def\expandafter\UrlBreaks\expandafter{\UrlBreaks
  \do\a\do\b\do\c\do\d\do\e\do\f\do\g\do\h\do\i\do\j%
  \do\k\do\l\do\m\do\n\do\o\do\p\do\q\do\r\do\s\do\t%
  \do\u\do\v\do\w\do\x\do\y\do\z\do\A\do\B\do\C\do\D%
  \do\E\do\F\do\G\do\H\do\I\do\J\do\K\do\L\do\M\do\N%
  \do\O\do\P\do\Q\do\R\do\S\do\T\do\U\do\V\do\W\do\X%
  \do\Y\do\Z}

\usepackage{algorithm}
\usepackage[noend]{algpseudocode}
\usepackage{algorithmicx}

\algrenewcommand\algorithmicindent{1.0em}

\algnewcommand\algorithmicforeach{\textbf{for each}}
\algdef{S}[FOR]{ForEach}[1]{\algorithmicforeach\ #1\ \algorithmicdo}

\algnewcommand\AlgAnd{\textbf{and} }
\algnewcommand\AlgOr{\textbf{or} }
\algnewcommand\AlgContinue{\textbf{Continue}}
\algnewcommand\AlgBreak{\textbf{break}}

\algrenewcommand\textproc{}

\algnewcommand{\Initialize}[1]{%
\State \textbf{Initialize:}%
\State \hspace*{\algorithmicindent}\parbox[t]{0.8\linewidth}{\raggedright #1}}

\algnewcommand{\LeftComment}[1]{\Statex $\triangleright$ #1 \hfill}

\algnewcommand{\IIf}[1]{\State\algorithmicif\ #1\ \algorithmicthen}
\algnewcommand{\EndIIf}{\unskip}

\newcommand{\PointODE}{PointODE}
\newcommand{\PointODENaive}{PointODE-Naive}
\newcommand{\PointODEElite}{PointODE-Elite}
\newcommand{\ResPBlock}{ResPBlock}
\newcommand{\ODEPBlock}{ODEPBlock}

\def\BibTeX{{\rm B\kern-.05em{\sc i\kern-.025em b}\kern-.08em
  T\kern-.1667em\lower.7ex\hbox{E}\kern-.125emX}}

\title{{\PointODE}: Lightweight Point Cloud Learning with Neural Ordinary Differential Equations on Edge}

\renewcommand{\shorttitle}{{\PointODE}: Lightweight Point Cloud Learning with Neural ODEs on Edge}

\hypersetup{
  pdftitle = {{\PointODE}: Lightweight Point Cloud Learning with Neural Ordinary Differential Equations on Edge},
  pdfsubject = {cs.LG},
  pdfauthor = {Keisuke Sugiura, Mizuki Yasuda, Hiroki Matsutani},
  pdfkeywords = {FPGA, Point Cloud, PointMLP, Residual Networks, Neural Ordinary Differential Equations}
}

\author{%
  \href{https://orcid.org/0000-0001-8534-2381}%
  {\includegraphics[scale=0.06]{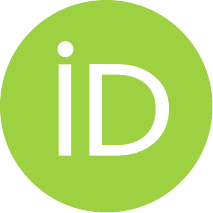}%
  \hspace{1mm}Keisuke Sugiura}\\
  University of Tsukuba\\
  1-1-1 Tenn\^{o}dai, Tsukuba, Ibaraki, Japan\\
  \texttt{sugiura@lila.cs.tsukuba.ac.jp}\\
  \And
  {Mizuki Yasuda}\\
  Keio University\\
  3-14-1 Hiyoshi, Kohoku-ku, Yokohama, Japan\\
  \texttt{yasuda@arc.ics.keio.ac.jp}\\
  \And
  \href{https://orcid.org/0000-0001-9578-3842}%
  {\includegraphics[scale=0.06]{orcid.pdf}%
  \hspace{1mm}Hiroki Matsutani}\\
  Keio University\\
  3-14-1 Hiyoshi, Kohoku-ku, Yokohama, Japan\\
  \texttt{matutani@arc.ics.keio.ac.jp}
}

\begin{document}

\maketitle



\begin{abstract}
Embedded edge devices are often used as a computing platform to run real-world point cloud applications, but recent deep learning-based methods may not fit on such devices due to limited resources.
In this paper, we aim to fill this gap by introducing {\PointODE}, a parameter-efficient ResNet-like architecture for point cloud feature extraction based on a stack of MLP blocks with residual connections.
We leverage Neural ODE (Ordinary Differential Equation), a continuous-depth version of ResNet originally developed for modeling the dynamics of continuous-time systems, to compress {\PointODE} by reusing the same parameters across MLP blocks.
The point-wise normalization is proposed for {\PointODE} to handle the non-uniform distribution of feature points.
We introduce {\PointODEElite} as a lightweight version with 0.58M trainable parameters and design its dedicated accelerator for embedded FPGAs.
The accelerator consists of a four-stage pipeline to parallelize the feature extraction for multiple points and stores the entire parameters on-chip to eliminate most of the off-chip data transfers.
Compared to the ARM Cortex-A53 CPU, the accelerator implemented on a Xilinx ZCU104 board speeds up the feature extraction by 4.9x, leading to 3.7x faster inference and 3.5x better energy-efficiency.
Despite the simple architecture, {\PointODEElite} shows competitive accuracy to the state-of-the-art models on both synthetic and real-world classification datasets, greatly improving the trade-off between accuracy and inference cost.
\end{abstract}



\section{Introduction} \label{sec:intro}
Point cloud is a collection of points representing 3D scenes.
Thanks to the increasing availability of low-cost 3D scanners (e.g., LiDARs and depth cameras), it serves as a basis for various mobile-edge applications including 3D reconstruction~\cite{SungjoonChoi15,JaesikPark17}, mapping~\cite{JiZhang14,TixiaoShan18,TixiaoShan20,HanWang21}, and object tracking~\cite{XingyiZhou20,TianweiYin21}.
Since PointNet~\cite{CharlesRQi17A} and PointNet++~\cite{CharlesRQi17B}, deep learning-based point cloud analysis has been the subject of extensive research and gained tremendous success.
PointNet is the first successful architecture for learning a global representation of point clouds via MLP-based point-wise feature extraction and global pooling.
PointNet++ adopts a hierarchical PointNet architecture to capture local geometric structures at varying scales.
The follow-up methods employ various techniques such as custom convolution operators~\cite{YiqunLin20,MutianXu21}, graph convolutions~\cite{HaoranZhou21,ZhiHaoLin22}, and transformers~\cite{XuminYu22,XianFengHan22} to enhance the feature representation, at the cost of increasing model complexity.

\begin{figure}[htbp]
  \centering
  \includegraphics[keepaspectratio, width=0.5\linewidth]{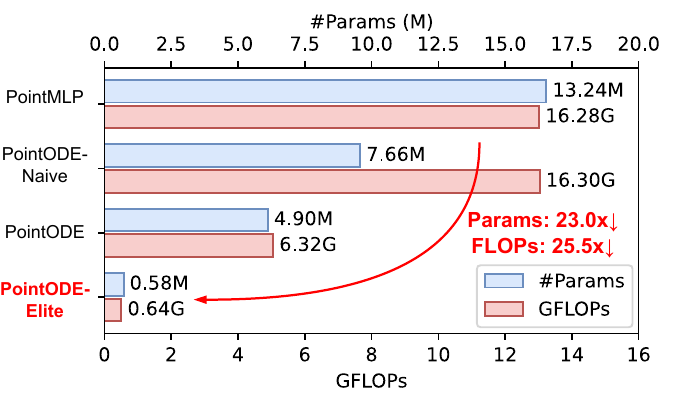}
  \caption{Number of parameters and FLOPs of the proposed {\PointODEElite} and the baseline PointMLP.}
  \label{fig:model-size}
\end{figure}

The recently-proposed PointMLP~\cite{XuMa22} takes the opposite direction and avoids the use of complicated feature extractors.
PointMLP is composed of a stack of MLP blocks with residual connections, but achieves competitive accuracy to state-of-the-art models on popular tasks such as classification and part segmentation.
In addition, PointMLP can efficiently leverage the computing power of FPGAs, as it mainly involves highly-parallelizable matrix operations within MLPs.
PointMLP is basically a ResNet~\cite{KaimingHe16}-like architecture with MLPs instead of 2D convolutions.
Compared to convolutions, FC (Fully-Connected) layers require a large number of parameters proportional to the feature dimensions.

Interestingly, ResNet can be interpreted as a discrete approximation of an ODE.
Based on this close relationship, ResNet is extended to a continuous-depth version dubbed as Neural ODE~\cite{RickyTQChen18}, to represent the hidden dynamics of continuous-time systems.
It is widely adopted in a variety of tasks, e.g., time-series forecasting~\cite{YuliaRubanova19,ZonghanWu20,PatrickKidger20}, trajectory modeling~\cite{YuxuanLiang21,SongWen22}, climate forecasting~\cite{JeehyunHwang21,YogeshVerma24}, physics simulation~\cite{GuangsiShi24}, video generation~\cite{SunghyunPark21,YuchengXu24}, and image registration~\cite{YifanWu22}.
Neural ODE is inherently more parameter-efficient than ResNet, as it is equivalent to a ResNet with an arbitrary number of residual blocks sharing the same parameters.
From this perspective, Neural ODE can be treated as a powerful model compression technique for ResNet-like models.
The resulting model consists of fewer residual blocks that are reused during inference, thereby reducing both logic and memory resource consumptions and making it well-suited to embedded FPGAs.
While the complicated design of recent feature extractors has limited the use of Neural ODE, it can be easily applied to PointMLP with minimal architectural changes, thus bridging the research gap between ODEs and point cloud DNNs.

In this paper, we introduce a new DNN architecture for point cloud analysis, \textbf{\PointODE}, combining the advantages of both PointMLP and Neural ODE.
{\PointODE} is a simple residual MLP-based network but integrates Neural ODE to effectively reduce the number of parameters by replacing a sequence of residual blocks with a single reusable block (Fig. \ref{fig:model-size}).
By adjusting the network size, we propose a lightweight version named \textbf{\PointODEElite} with 0.58M learnable parameters, significantly improving the accuracy-computation trade-off.

We then design a dedicated accelerator for {\PointODEElite}, which incorporates a set of MLP blocks and the Euler numerical integration method for solving ODEs.
{\PointODE} omits the geometric affine transformation in PointMLP and instead performs a point-wise normalization to avoid the necessity of computing global statistics of input feature points, leading to the improved parallelism and accuracy.
The FPGA design consists of a four-stage feature extraction pipeline to exploit the point-level parallelism.
Thanks to the lower parameter size and reuse of blocks, parameters and intermediate activations of the entire model fit on the FPGA memory, eliminating most of the off-chip data transfers.
The proposed accelerator is implemented on a Xilinx ZCU104 board and evaluated on classification datasets.
It runs 4.9x faster than the ARM Cortex-A53 processor, while achieving a comparable accuracy to state-of-the-art networks.
The contributions of this paper are summarized as follows:
\begin{itemize}
  \item We propose {\PointODE} and its lightweight counterpart {\PointODEElite} as a simple yet highly-accurate DNN architecture for point clouds.
  \item To the best of our knowledge, this paper is the first to explore an FPGA design for the Neural ODE-based point cloud analysis.
\end{itemize}


\section{Related Work} \label{sec:related}
\textbf{DNNs for Point Clouds}:
PointNet~\cite{CharlesRQi17A} is a pioneering model that directly consumes point clouds.
It extracts point-wise features using a shared MLP and aggregates them into a global representation via a symmetric pooling function.
PointNet++~\cite{CharlesRQi17B} is an extension that aims to capture fine geometric context through repeated sampling and grouping of feature points.
PointNet and PointNet++ have catalyzed the development of more sophisticated networks to learn better feature representations at the cost of increased complexity.

One approach is to design custom convolution kernels for point clouds~\cite{MatanAtzmon18,YifanXu18,YangyanLi18,YongchengLiu19,WenxuanWu19,YiqunLin20,MutianXu21}, while several works~\cite{YueWang19,QiangengXu20,HuanLei21,HaoranZhou21,ZhiHaoLin22} employ GCNs (Graph Convolutional Networks) to process KNN graphs built from point clouds.
Inspired by the tremendous success, transformer-based methods~\cite{MengHaoGuo21,HengshuangZhao21,XuminYu22,XianFengHan22} employ a self-attention mechanism to capture the relationship between all points, while it incurs a quadratic computational cost in terms of input size.
Another approach is to represent point clouds as curves (sequence of points)~\cite{TiangeXiang21}, polygon meshes~\cite{YutongFeng19}, or umbrella surfaces~\cite{HaoxiRan22}.
Aside from these approaches, PointNeXt~\cite{GuochengQian22} revisit PointNet++ and proposes improved training strategies to boost its performance without changing the network structure.
PointMLP~\cite{XuMa22} opts to use a stack of MLP blocks with residual connections and learns a hierarchical feature representation like PointNet++.
PointMLP achieves competitive accuracy and faster inference speed compared to the existing complex feature extractors, demonstrating the effectiveness of a simple MLP-based architecture.

\textbf{ODE-based DNNs}:
Neural ODE~\cite{RickyTQChen18} is initially developed as a method to model the hidden dynamics (i.e., derivatives) of a continuous-time system using a neural network.
Various follow-up works have emerged to e.g., enhance the expressive power~\cite{EmilienDupont19}, obtain higher-order derivatives~\cite{GuanHorngLiu21}, avoid incorrect gradients and numerical instability~\cite{AmirGholami19,TianjunZhang19}, improve robustness to noise~\cite{XuanqingLiu19}, and handle graph data~\cite{MichaelPoli19,LouisPascalXhonneux20}.
ExNODE~\cite{YangLi20} is a seminal work that extends Neural ODE to point clouds.
It uses DeepSet~\cite{ManzilZaheer17} or Set Transformer~\cite{JuhoLee19} as a basic component to ensure the permutation invariance.
CaSPR~\cite{DavisRempe20} uses Neural ODE to learn the representations of dynamically moving point clouds.
Exploiting the connection between ResNets~\cite{KaimingHe16} and Neural ODEs, the recent work~\cite{HirokiKawakami23,IkumiOkubo24} utilizes Neural ODE as a technique to compress ResNet-like image classification models without sacrificing accuracy.
Taking a similar approach, this work integrates Neural ODE into PointMLP and builds a lightweight network to perform point cloud analysis on embedded devices.

\textbf{Neural ODEs on FPGAs}:
Only a few works~\cite{LeiCai23,HirokiKawakami23,YiChen23,IkumiOkubo24} explore the FPGA design of Neural ODEs.
The authors of \cite{HirokiKawakami23} combine Neural ODE and separable convolution to develop a lightweight network with only 0.6M parameters, which is implemented on a Xilinx ZCU104 board.
In \cite{IkumiOkubo24}, an FPGA accelerator is proposed for a CNN-Transformer hybrid model combining Neural ODE and ViT (Vision Transformer)~\cite{AlexeyDosovitskiy21}.
In \cite{LeiCai23}, the third-order Runge-Kutta method is used as an ODE solver instead of the first-order Euler method, while this requires 3x more network forward passes per iteration.
A two-stage structured pruning method along with a history-based step size search is devised to mitigate this problem.
While these accelerators achieve favourable performance, they are only evaluated on small-scale image datasets (e.g., CIFAR-10) and not designed for 3D point clouds.


\section{Background} \label{sec:prelim}
\subsection{Neural ODE} \label{sec:prelim-neural-ode}
\begin{figure}[htbp]
  \centering
  \includegraphics[keepaspectratio, width=0.55\linewidth]{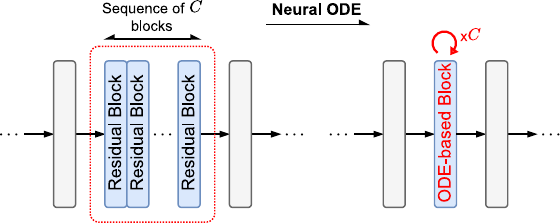}
  \caption{Replacing the forward pass of $C$ ResBlocks with $C$ forward iterations of the single ODE-based building block.}
  \label{fig:neural-ode}
\end{figure}

ResNet~\cite{KaimingHe16} employs residual (skip, shortcut) connections in its building blocks to stabilize the training of deep networks.
A sequence of buliding blocks (\textbf{ResBlock}s) can be expressed as the following recursive formula:
\begin{equation}
  \vb{h}_t = \vb{h}_{t - 1} + f(\vb{h}_{t - 1}, \vb*{\theta}_t),
  \label{eq:resblock}
\end{equation}
where $t = 1, \ldots$ denotes a block index, $\vb{h}_{t - 1}$ an input, $f$ a stack of layers, and $\vb*{\theta}_t$ a set of trainable parameters in a block.
The addition represents a residual connection.
Eq. \ref{eq:resblock} can be viewed as a forward Euler discretization of the ODE $\dd{\vb{h}(t)} / \dd{t} = f(\vb{h}(t), t, \vb*{\theta})$, where $t$ denotes a continuous time and $f(\cdot, \vb*{\theta})$ represents a time-derivative (i.e., dynamics) of a hidden state $\vb{h}(t)$.
Neural ODE~\cite{RickyTQChen18} formulates the forward propagation of ResBlocks as a solution of this ODE.
Given an input $\vb{h}(t_a)$, an output $\vb{h}(t_b)$ is obtained using an arbitrary ODE solver (i.e., by integrating $f$ over $[t_a, t_b]$).
For instance, the Euler method can be used:
\begin{equation}
  \vb{h}(t_j) = \vb{h}(t_{j - 1}) + h f(\vb{h}(t_{j - 1}), t_{j - 1}, \vb*{\theta}),
  \label{eq:euler}
\end{equation}
where $j = 0, \ldots, C$ denotes a discrete time step ($t_j = t_a + jh$, $h = (t_b - t_a) / C$) and $C$ is the number of iterations.
Eq. \ref{eq:euler} represents an ODE-based building block (\textbf{ODEBlock}) consisting of layers $f$, parameters $\vb*{\theta}$, and a residual connection.
As depicted in Fig. \ref{fig:neural-ode}, ODEBlock uses the same network $f$ during $C$ forward passes, which is $C$ times more parameter-efficient than using $C$ separate ResBlocks.
Neural ODE can be seen as a technique to compress residual networks by fusing multiple similar ResBlocks into a single ODEBlock.

\subsection{PointMLP} \label{sec:prelim-pointmlp}
\begin{figure}[htbp]
  \centering
  \includegraphics[keepaspectratio, width=0.6\linewidth]{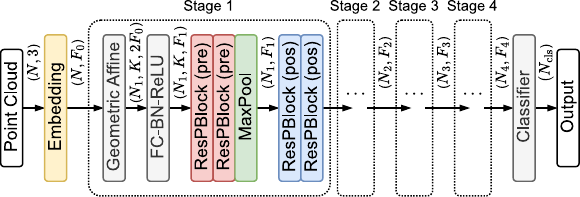}
  \caption{Architecture of PointMLP.}
  \label{fig:pointmlp}
\end{figure}

PointMLP~\cite{XuMa22} is a pure MLP-based network with residual connections for point cloud feature extraction.
The architecture is presented in Fig. \ref{fig:pointmlp}.
PointMLP directly takes 3D point coordinates $\mathcal{P} = \{ \vb{p}_1, \ldots, \vb{p}_N \}$ as input and extracts high-dimensional features for a subset of points, which are fed to the network for a specific task (e.g., classification and segmentation).
PointMLP consists of an embedding block to extract $F_0$-dim local features $\{ \vb{f}_1^0, \ldots, \vb{f}_N^0 \}$ for each point, which is followed by a stack of four stages to hierarchically aggregate these local features.

Each stage $s \in [1, 4]$ produces $F_s$-dim features $\mathcal{F}^s = \{ \vb{f}_i^s \}$ for $N_s$ sampled points $\mathcal{P}^s = \{ \vb{p}_i^s \}$.
PointMLP first samples $N_s$ points (i.e., group centroids) from $\mathcal{P}^{s - 1}$ and finds their $K$-nearest neighbors (NNs).
For each group centroid $\vb{p}_j^{s - 1} \in \mathcal{P}^{s - 1}$, the geometric affine module applies an affine transformation to the features $\{ \vb{f}_{j, k}^{s - 1} \}$ of its neighbors $\{ \vb{p}_{j, k}^{s - 1} \}$.
The transformed features $\{ \hat{\vb{f}}_{j, k}^{s - 1} \}$ of grouped points are then concatenated with that of the centroid $\vb{f}_j^{s - 1} \in \mathcal{F}^{s - 1}$, producing a tensor of size $(N_s, K, 2F_{s - 1})$.
As shown in Fig. \ref{fig:pointmlp} (\textbf{FC-BN-ReLU} and \textbf{ResPBlock (pre)}), an MLP block and the first two residual point blocks (\textbf{ResPBlock}s) transform these $2 F_{s - 1}$-dim features into $F_s$-dim features and produce an output of size $(N_s, K, F_s)$.
The max-pooling aggregates features of $K$ local neighbors into a single $F_s$-dim feature that capture the geometry of the whole group.
The subsequent two ResPBlocks (\textbf{ResPBlock (pos)} in Fig. \ref{fig:pointmlp}) extract deeply aggregated features and produces an output $\mathcal{F}^s$ of size $(N_s, F_s)$.

PointMLP operates on raw point clouds instead of 3D grids or 2D images and hence does not require costly preprocessing (e.g., voxelization and multi-view rendering).
Besides, it does not rely on sophisticated feature extractors (e.g., custom convolution operators~\cite{YangyanLi18,WenxuanWu19,HuguesThomas19,MutianXu21}).
ResPBlock (Fig. \ref{fig:respblock-odepblock}, left) has a simple architecture consisting of FC layers along with BN (Batch Normalization) and ReLU.
It offers high parallelism as the feature extraction is performed independently for each point.
PointMLP is thus amenable to FPGA acceleration thanks to its simple architecture and high parallelism.


\section{{\PointODEElite}} \label{sec:method}
This section describes the architecture of {\PointODEElite}, which is built upon PointMLP.
For better parameter-efficiency and parallelism, {\PointODEElite} introduces the three improvements: (1) ODE-based residual point block, (2) residual block reordering, and (3) point-wise normalization.

\subsection{\ODEPBlock: ODE-based Residual Point Block} \label{sec:method-ode-block}
\begin{figure}[htbp]
  \centering
  \includegraphics[keepaspectratio, width=0.6\linewidth]{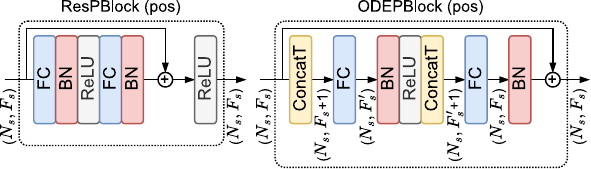}
  \caption{{\ResPBlock} and {\ODEPBlock}.}
  \label{fig:respblock-odepblock}
\end{figure}

Each stage in PointMLP has two sets of two {\ResPBlock}s for extracting local and aggregated features (Fig. \ref{fig:pointmlp}).
Using Neural ODE, a single forward pass of two consecutive {\ResPBlock}s can be replaced by two forward iterations of a single ODE-based residual point block (\textbf{\ODEPBlock}).
Since {\ResPBlock}s dominate the model size of PointMLP (i.e., account for 88.9\% of the total number of parameters), this change leads to a 1.73x parameter reduction (13.24M to 7.66M).
We refer to this model as \textbf{\PointODENaive}.
Note the number of forward iterations $C$ allows to balance the trade-off between accuracy and inference time (Fig. \ref{fig:acc-vs-iter}).
As depicted in Fig. \ref{fig:respblock-odepblock}, {\ODEPBlock} is similar to {\ResPBlock} but has two additional layers (\textbf{ConcatT}) to fuse the time information (i.e., a continuous time variable $t$) into the input features.

Formally, {\ODEPBlock} in stage $s$ takes $F_s$-dim features for $N_s$ points and produces an output of the same size.
The first ConcatT layer concatenates a time variable $t$ to the input, creating a set of $(F_s + 1)$-dim features, which is passed to an FC-BN-ReLU block to extract $F_s'$-dim features.
Followed by another ConcatT layer, an FC-BN block transforms $(F_s' + 1)$-dim features into $F_s$-dim ones, which are then added to the original input via a residual connection.

\subsection{Residual Block Reordering} \label{sec:method-reordering}
\begin{figure}[htbp]
  \centering
  \includegraphics[keepaspectratio, width=0.6\linewidth]{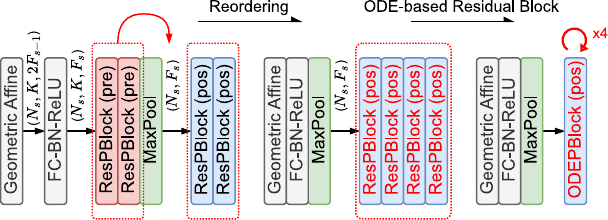}
  \caption{Residual block reordering (left: PointMLP, center: reordered layers, right: {\PointODE}(-Elite)).}
  \label{fig:pointmlp-reorder}
\end{figure}

\begin{figure}[htbp]
  \centering
  \includegraphics[keepaspectratio, width=0.55\linewidth]{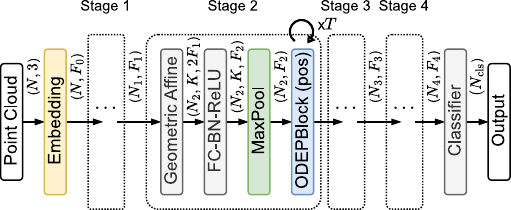}
  \caption{Architecture of {\PointODE} and {\PointODEElite}.}
  \label{fig:pointode}
\end{figure}

To improve the effectiveness of using ODE blocks, we propose to first reorder the blocks in each stage of PointMLP before employing Neural ODE.
As shown in Fig. \ref{fig:pointmlp-reorder}, the first two {\ResPBlock}s are moved next to the max-pooling layer, such that four {\ResPBlock}s (now stacked in sequence) can be replaced by four forward iterations of the same {\ODEPBlock}.
This model has 1.56x fewer parameters than {\PointODENaive} (7.66M to 4.90M), as each stage has only one {\ODEPBlock} instead of two\footnote{The first stage keeps two separate {\ODEPBlock}s for accuracy.}.
The resulting model architecture is shown in Fig. \ref{fig:pointode}.
An MLP block (FC-BN-ReLU) extracts $F_s$-dim local features for each of $K$ neighboring points of a group centroid, producing an output of size $(N_s, K, F_s)$, while an ODEPBlock learns an aggregated feature for each group.
Following PointMLP, $F_s$ and $N_s$ are set to $2F_{s - 1}$ and $N_{s - 1} / 2$, respectively, such that each stage extracts higher-level features for fewer representative points.

\subsection{Point-wise Normalization} \label{sec:method-local-norm}
In each stage $s$ of PointMLP, the geometric affine module first transforms features $\{ \vb{f}_{j, k}^{s - 1} \}$ of the $K$ neighborhood points $\{ \vb{p}_{j, k}^{s - 1} \}$ in a group as follows:
\begin{equation}
  \hat{\vb{f}}_{j, k}^{s - 1} = \vb*{\alpha} \odot \frac{\vb{\Delta f}_{j, k}^{s - 1}}{\sigma + \varepsilon} + \vb*{\beta}, \quad
  \vb{\Delta f}_{j, k}^{s - 1} = \vb{f}_{j, k}^{s - 1} - \vb{f}_j^{s - 1},
  \label{eq:pointmlp-geometric-affine}
\end{equation}
where $\vb{f}_j^{s - 1} \in \mathcal{F}^{s - 1}$ is an $F_{s - 1}$-dim feature associated with the group centroid $\vb{p}_j^{s - 1} \in \mathcal{P}^{s - 1}$.
The two $F_{s - 1}$-dim vectors $\vb*{\alpha}$ and $\vb*{\beta}$ are trainable scale and offset parameters, respectively, and $\varepsilon = \text{1e-5}$ is a small constant to avoid zero division.
The scalar $\sigma$ is defined as:
\begin{equation}
  \sigma = \sqrt{\frac{1}{N_{s - 1} K} \sum_{j, k}
    \left\| \vb{\Delta f}_{j, k}^{s - 1} - \vb*{\mu} \right\|_2^2}, \quad
  \vb*{\mu} = \frac{1}{N_{s - 1} K} \sum_{j, k} \vb{\Delta f}_{j, k}^{s - 1}.
  \label{eq:pointmlp-geometric-affine-sigma}
\end{equation}
Since $\vb*{\mu}$ and $\sigma$ are the global mean and standard deviation of the feature residual $\vb{\Delta f}_{j, k}^{s - 1}$ across all $N_{s - 1}$ points, the affine transform $\vb{f} \mapsto \hat{\vb{f}}$ is not fully independent for each point.
Each stage cannot proceed to the normalization (Eq. \ref{eq:pointmlp-geometric-affine}) and subsequent blocks until $\sigma$ is available.
Besides, the accumulation of $N_{s - 1} K$ features requires a wider fixed-point format to avoid overflow
Considering these, we propose to compute $\mu$ and $\sigma$ separately for each point in a group (similar to layer normalization~\cite{JimmyLeiBa16}) and transform the features $\{ \vb{f}_{j, k}^{s - 1} \}$ as follows:
\begin{equation}
  \tilde{\vb{f}}_{j, k}^{s - 1} = \vb*{\alpha} \odot \frac{\vb{\Delta f}_{j, k}^{s - 1}}{\tilde{\sigma}_{j, k} + \varepsilon} + \vb*{\beta},
  \label{eq:pointmlp-layer-norm}
\end{equation}
\begin{equation}
  \tilde{\sigma}_{j, k} = \sqrt{\frac{1}{F_{s - 1}}
    \left\| \vb{\Delta f}_{j, k}^{s - 1} - \tilde{\mu}_{j, k} \vb{1} \right\|_2^2}, \quad
  \tilde{\mu}_{j, k} = \frac{1}{F_{s - 1}} \vb{1}^\top \vb{\Delta f}_{j, k}^{s - 1},
  \label{eq:pointmlp-layer-norm-sigma}
\end{equation}
where $\vb{1}$ is an $F_{s - 1}$-dim vector of ones.
In this case, $\tilde{\mu}$ and $\tilde{\sigma}$ represent the mean and standard deviation of the elements in $\vb{\Delta f}_{j, k}^{s - 1}$.
The affine transform (Eq. \ref{eq:pointmlp-layer-norm}) can be performed independently for each point, which allows for a pipelined execution of blocks in a stage (Sec. \ref{sec:impl}).
In addition, such normalization can better handle irregular point clouds with a varying point distribution by adjusting the scaling factor $\tilde{\sigma}_{j, k}$ for each point $\vb{p}_{j, k}$ in a group.
We experimentally validate the accuracy improvements with point-wise normalization in Sec. \ref{sec:eval}.
We refer to the model with block reordering (Sec. \ref{sec:method-reordering}) and improved normalization as \textbf{\PointODE}.

\subsection{{\PointODEElite}: Lightweight Version of {\PointODE}} \label{sec:method-elite}
Following PointMLP-Elite~\cite{XuMa22}, \textbf{\PointODEElite} is introduced as a lightweight version that makes three modifications to {\PointODE}.
The feature dimensions $F_1, \ldots, F_4$ are reduced from $(128, 256, 512, 1024)$ to $(64, 128, 256, 256)$.
{\ODEPBlock}s employ the bottleneck structure by setting the number of intermediate feature dimensions to $F_s' = F_s / 4$ instead of $F_s' = F_s$ as in {\PointODE} (Fig. \ref{fig:respblock-odepblock}).
The output dimensions of the embedding block $F_0$ and the group size $K$ are both halved from 64 to 32, and from 24 to 12, respectively (Fig. \ref{fig:pointode}).
With these changes, {\PointODEElite} achieves a 8.52x and 9.90x reduction of the parameter size and FLOPs (4.90M to 0.58M, 6.32G to 0.64G), respectively, leading to a 23.02x and 25.51x overall reduction than the baseline PointMLP.
Notably, the feature extraction part (the embedding block and four stages) only has 0.30M parameters.
Even compared to PointMLP-Elite, {\PointODEElite} has 1.25x and 1.83x fewer parameters and FLOPs, while achieving the same or slightly higher accuracy.

Considering the unordered nature of point clouds, the network output should be invariant to the ordering of input points~\cite{CharlesRQi17A}.
ExNODE~\cite{YangLi20} proves that the output of Neural ODE ($\vb{h}(t_C)$ in Eq. \ref{eq:euler}) is permutation invariant given the input feature set $\vb{h}(t_0)$, if the network $f(\vb{h}(t), t, \vb*{\theta})$ is permutation invariant with respect to $\vb{h}(t)$.
{\ODEPBlock} satisfies this property, because the network layers (i.e., ConcatT, FC, BN and ReLU) apply their respective operations to each $F_s$-dim point feature independently.
The other components (i.e., point-wise normalization, FC-BN-ReLU, and max-pooling) are not affected by the ordering of inputs as well, indicating that {\PointODE} produces permutation invariant features.


\section{FPGA Design and Implementation} \label{sec:impl}
In this section, we describe the FPGA design and implementation of {\PointODEElite}.
We offload the feature extraction part (the embedding block and four stages) to the FPGA, as it accounts for $>$90\% of the entire inference time (Fig. \ref{fig:time-bar}).

\subsection{Point-wise Normalization} \label{sec:impl-norm}
\begin{figure}[htbp]
  \centering
  \includegraphics[keepaspectratio, width=0.6\linewidth]{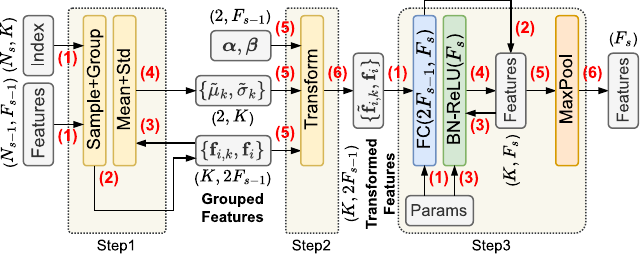}
  \caption{Block diagram of modules for the point-wise normalization, MLP block (FC-BN-ReLU), and max-pooling.}
  \label{fig:impl-steps1-3}
\end{figure}

Fig. \ref{fig:impl-steps1-3} illustrates the block diagram of a stage (except {\ODEPBlock}).
The $F_{s - 1}$-dim features $\mathcal{F}^{s - 1}$ for $N_{s - 1}$ points along with the indices of size $(N_s, K)$ are stored on the respective buffers and passed as input to the stage.
First, the \textbf{Sample+Group} module reads a single row of the index buffer.
Based on this, it accesses the feature buffer to collect a feature $\vb{f}_j^{s - 1} \in \mathcal{F}^{s - 1}$ of a group centroid as well as those of the $K$ neighboring points $\{ \vb{f}_{j, k}^{s - 1} \}$ belonging to the same group.
The concatenated features $\{ [\vb{f}_{j, k}^{s - 1}, \vb{f}_j^{s - 1}] \}$ are stored on an intermediate buffer of size $(K, 2F_{s - 1})$.
Then, the \textbf{Mean+Std} module computes a point-wise mean $\{ \tilde{\mu}_{j, k} \}$ and deviation $\{ \tilde{\sigma}_{j, k} \}$ (Eq. \ref{eq:pointmlp-layer-norm-sigma}) and writes them to another buffer of size $(2, K)$.

The \textbf{Transform} module normalizes grouped features (Eq. \ref{eq:pointmlp-layer-norm}) and the result $\{ [\tilde{\vb{f}}_{j, k}^{s - 1}, \vb{f}_j^{s - 1}] \}$ is fed to an MLP block consisting of \textbf{FC} and \textbf{BN-ReLU} modules.
Given an input feature $\vb{x}$, a weight matrix $\vb{W}$, and a bias vector $\vb{b}$ stored on-chip, the \textbf{FC} module performs a dot product and an addition $\vb{y} = \vb{W} \vb{x} + \vb{b}$.
BN and ReLU nonlinearity are fused together into the \textbf{BN-ReLU} module, which computes an output $\vb{y} = \max(\vb*{\gamma} \vb{x} + \vb*{\delta}, \vb{0})$ based on a BN scale $\vb*{\gamma}$ and bias $\vb*{\delta}$ stored on-chip.
The \textbf{MaxPool} module retrieves $F_s$-dim features of the $K$ grouped points from \textbf{BN-ReLU} and aggregates them into a single $F_s$-dim feature via max-pooling, which is stored on an output buffer (Fig. \ref{fig:impl-steps1-3}, right).
DSP blocks are utilized to compute the statistics $\tilde{\mu}, \tilde{\sigma}$ in \textbf{Sample+Group}, normalize the grouped features in \textbf{Transform}, and perform MAC operations in \textbf{FC} as well as \textbf{BN-ReLU}.

The above process should be repeated $N_s$ times to extract features for all sampled points in $\mathcal{P}^s$.
Since each module processes a single group at a time, the intermediate buffers do not need to store features for the whole point cloud $\mathcal{P}^s$, thereby significantly reducing the on-chip memory usage\footnote{For instance, the buffer size for normalized features are reduced from $(N_s, K, 2F_{s - 1})$ to just $(K, 2F_{s - 1})$ by processing a single sampled point (group centroid) at a time.}.

\subsection{{\ODEPBlock}} \label{sec:impl-odepblock}
\begin{figure}[htbp]
  \centering
  \includegraphics[keepaspectratio, width=0.6\linewidth]{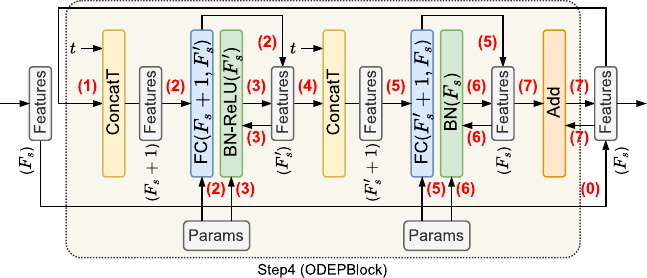}
  \caption{Block diagram of {\ODEPBlock}.}
  \label{fig:impl-step4}
\end{figure}

Fig. \ref{fig:impl-step4} shows the block diagram of modules for {\ODEPBlock} at stage $s$.
{\ODEPBlock} is realized by a set of modules corresponding to the layers and a residual connection (Fig. \ref{fig:respblock-odepblock}, right), as well as on-chip buffers for storing layer parameters and intermediate results.
An $F_s$-dim aggregated feature is first read from the on-chip buffer and the output buffer is initialized (Fig. \ref{fig:impl-step4}, right).
The ODE forward pass (Eq. \ref{eq:euler}) is then repeated for $C$ iterations to gradually update the $F_s$-dim output feature.
The \textbf{ConcatT} module combines a time variable $t \in [t_a, t_b]$ with an input to produce an $(F_s + 1)$-dim augmented feature, which is projected to $F_s'$-dim by the \textbf{FC} and \textbf{BN-ReLU} modules.
The intermediate $F_s'$-dim feature is processed by a sequence of \textbf{ConcatT}-\textbf{FC}-\textbf{BN} modules to obtain an $F_s$-dim output, which is scaled by the step size $h$ and then added to the original input following the Euler numerical integration method (Eq. \ref{eq:euler}).
The operation of each module is parallelized by unrolling the loop and partitioning the buffers as needed.
Similar to Sec. \ref{sec:impl-norm}, the sampled points $\mathcal{P}^s$ (groups) are processed one at a time, which greatly saves the on-chip memory.

\subsection{Implementation of the Stage} \label{sec:impl-stage}
\begin{figure}[htbp]
  \centering
  \includegraphics[keepaspectratio, width=0.6\linewidth]{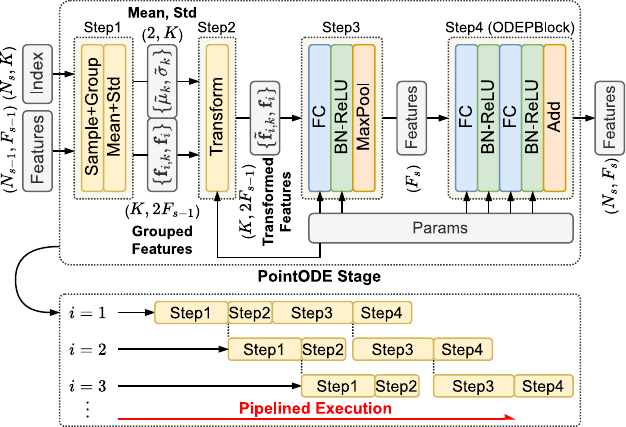}
  \caption{Block diagram of the stage.}
  \label{fig:impl-stage}
\end{figure}

The overall architecture of the stage $s$ is depicted in Fig. \ref{fig:impl-stage}.
The modules presented in Sec. \ref{sec:impl-norm}--\ref{sec:impl-odepblock} are organized as a four-step pipeline (Fig. \ref{fig:impl-stage}, bottom), where each step processes a different sampled point in $\mathcal{P}^s$ and its $K$ nearest neighbors.
The first step is formed by \textbf{Sample+Group} and \textbf{Mean+Std} modules, while \textbf{Transform} module works as the second step (Fig. \ref{fig:impl-steps1-3}).
The \textbf{FC}, \textbf{BN-ReLU}, and \textbf{MaxPool} modules are tied together to form the third step (Fig. \ref{fig:impl-steps1-3}), whereas the last step consists of an {\ODEPBlock} (Fig. \ref{fig:impl-step4}).
Since {\PointODEElite} incorporates the proposed point-wise normalization (Eqs. \ref{eq:pointmlp-layer-norm}--\ref{eq:pointmlp-layer-norm-sigma}) instead of the geometric affine module (Eqs. \ref{eq:pointmlp-geometric-affine}--\ref{eq:pointmlp-geometric-affine-sigma}), each sampled point can be processed independently.
Unlike PointMLP, each stage can start running the subsequent MLP and {\ODEPBlock} without waiting for the computation of $\sigma$, which allows for pipelined execution.
By exploiting the parallelism of {\PointODEElite} and processing four points concurrently, the total latency of the stage is reduced by 2.83x.

\subsection{Implementation of {\PointODEElite}} \label{sec:impl-pointode-elite}
\begin{figure}[htbp]
  \centering
  \includegraphics[keepaspectratio, width=0.6\linewidth]{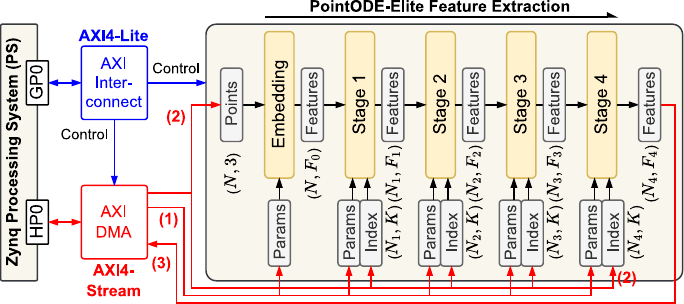}
  \caption{Overview of the FPGA implementation.}
  \label{fig:impl-overview}
\end{figure}

Fig. \ref{fig:impl-overview} shows an overview of the FPGA implementation.
{\PointODEElite} is directly mapped onto the FPGA fabric, and the implementation can be divided into four stages as well as an embedding block, with each consisting of a set of modules.
The embedding block (Fig. \ref{fig:pointode}) extracts $F_0$-dim features for each of $N$ input points using the \textbf{FC} and \textbf{BN-ReLU} modules and hands these features off to the first stage.

{\PointODEElite} is designed as an IP core with two 32-bit AXI interfaces.
It uses an AXI4-Stream interface to stream in the input point clouds, parameters (for FC, BN, and normalization), and sampling indices, as well as to stream out the extracted features to DDR.
The DMA controller handles the data movement between DDR and {\PointODEElite}.
Besides, the host writes configuration parameters of {\PointODEElite} and the DMA controller (e.g., the number of ODE iterations $C$ and the ODE step size $h$) through an AXI4-Lite interface.
These AXI interfaces are connected to the Processing System (PS) side via high-performance ports.

The whole {\PointODEElite} and intermediate buffers fit in the on-chip memory of embedded FPGAs (e.g., ZCU104) thanks to the parameter-efficiency of the network and coarse-grained pipelining within each stage, which substantially reduces the data transfer overhead during inference.
The parameters are read from DDR and stored to the respective on-chip buffers (Fig. \ref{fig:impl-overview}) before inference.
At inference, the point cloud $\mathcal{P}$ as well as the sampling indices (of size $(N_s, K)$) are transferred to the on-chip buffers and fed forward to the {\PointODEElite} modules.
The extracted features (of size $(N_4, F_4)$) are written back to DDR via DMA and used for the specific tasks (e.g., classification).
As discussed in Sec. \ref{sec:impl-norm}, each stage $s$ needs precomputed indices of size $(N_s, K)$ to subsample $N_s$ points $\mathcal{P}^s$ out of $N_{s - 1}$ points $\mathcal{P}^{s - 1}$ and find $K$NNs for each sampled point.
The host performs farthest point sampling (FPS) and $K$NN search recursively on the input $\mathcal{P}$ four times to generate indices for $\mathcal{P}, \mathcal{P}_1, \mathcal{P}_2, \mathcal{P}_3$.
In addition to progressive downsampling, the classification network is also executed on the host after receiving extracted features.


\section{Evaluation} \label{sec:eval}
Here we evaluate the performance of {\PointODEElite} in terms of (i) accuracy, (ii) inference time, (iii) power consumption, and (iv) resource utilization.

\subsection{Experimental Setup} \label{sec:eval-setup}
\subsubsection{Implementation Details} \label{sec:eval-impl-details}
The proposed design (Fig. \ref{fig:impl-overview}) is implemented on a Xilinx ZCU104 board.
It features 2GB of DDR4 memory and an FPGA device (xczu7ev-2ffvc1156) containing a quad-core ARM Cortex-A53 processor clocked at 1.2GHz.
The Pynq (Python productivity for Zynq) framework is used to write the host code that interacts with the FPGA kernel.
We implement {\PointODEElite} using Vitis HLS 2020.2 and then invoke Vivado 2020.2 to generate a bitstream from the block design (Fig. \ref{fig:impl-overview}).
The clock frequency of {\PointODEElite} and the DMA controller is set to 200MHz.
We utilize a 24-bit fixed-point format with an 8-bit integer and a 16-bit fractional part to represent point clouds, extracted features, as well as parameters.
{\PointODEElite} handles the conversion between fixed- and floating-point representations.
We apply dataflow optimization (provided by Vitis HLS) to each stage for block-level pipelining (Sec. \ref{sec:impl-stage}).

\subsubsection{Point Cloud Datasets} \label{sec:eval-datasets}
We utilize two popular datasets for point cloud classification: ModelNet40~\cite{ZhirongWu15} and ScanObjectNN~\cite{MikaelaAngelinaUy19}.
ModelNet40 consists of 12311 point clouds across 40 categories (e.g., airplane, table), generated by uniformly sampling the surfaces of synthetic CAD objects.
The dataset is split into training and test sets with each containing 9843 and 2468 samples, respectively.
Each sample is translated and normalized to fit within a unit sphere centered at the origin.
ScanObjectNN is a real-world object dataset consisting of 2902 point clouds across 15 categories (e.g., desk, sofa).
Unlike ModelNet40, ScanObjectNN is more challenging as its samples are affected by various factors such as measurement error, background noise, and occlusion.
We subsample $N = 1024$ points out of 2048.
During training, we apply a random scaling of $[0.67, 1.5]$ and a random translation of $[-0.2, 0.2]$ along each axis to point clouds for data augmentation.

\subsubsection{Model Training Details} \label{sec:eval-train-details}
We train the networks using PyTorch on an Ubuntu 20.04 workstation equipped with Nvidia GeForce RTX 3090 GPUs.
The learning rate $\eta$ is initialized to 0.1 and 0.01 for ModelNet40 and ScanObjectNN, respectively\footnote{We set to $\eta = 0.02$ when training {\PointODEElite} on ScanObjectNN.}.
At the training phase, we adopt gradient clipping, a cross-entropy loss, an Adam optimizer with default parameters, and a cosine annealing scheduler with a minimum learning rate of 5e-3.
The networks are trained for 300 and 200 epochs with a batch size of 32 and 8 on ModelNet40 and ScanObjectNN, respectively.
In Neural ODE (Sec. \ref{sec:prelim-neural-ode}), the step size $h$ depends on the number of iterations $C$ as well as the integration interval $[t_a, t_b]$.
The initial time $t_a$ is fixed at 0, while the final time $t_b$ is chosen from $\{ 0.1, 0.2, 0.3 \}$ for each model and dataset to achieve the best accuracy.

\subsection{Classification Accuracy} \label{sec:eval-accuracy}
Table \ref{tbl:accuracy} summarizes the classification accuracy (with $N = 1024$ input points), where \textbf{mAcc} and \textbf{OA} denote the mean class accuracy and overall accuracy\footnote{The asterisk denotes surface normals are used along with point coordinates.}.
The results for PointMLP and {\PointODE} are obtained on the GPU workstation.
By replacing two consecutive ResPBlocks with one {\ODEPBlock}, {\PointODENaive} (Sec. \ref{sec:method-ode-block}) reduces the parameter size by 1.73x while keeping the accuracy within 0.6--1.6\% of PointMLP.
Compared to PointMLP, {\PointODEElite} consumes 23.02x less parameters and only shows an accuracy drop of 0.1--1.1\%, significantly improving a trade-off between model size and performance on both synthetic and real-world datasets.
It achieves on-par or even better accuracy with 1.25x fewer parameters than PointMLP-Elite.

While {\PointODEElite} is a simple network with residual MLPs and does not require normal information, it outperforms more sophisticated networks with custom convolution kernels~\cite{YangyanLi18,WenxuanWu19,HuguesThomas19}, graph convolution~\cite{YueWang19}, and transformers~\cite{MengHaoGuo21,HengshuangZhao21}.
{\PointODEElite} surpasses the existing ODE-based method (ExNODE~\cite{YangLi20} with Set Transformer~\cite{JuhoLee19}) by 4.1\% on ModelNet40, highlighting the effectiveness of hierarchical feature learning.
In addition, it shows competitive performance against state-of-the-art models~\cite{HaoxiRan22,GuochengQian22} with $>$4x fewer parameters.
These results demonstrate the effectiveness of Neural ODE-based approach for designing a parameter-efficient network for point clouds.

\begin{table}[htbp]
  \centering
  \caption{Classification accuracy.}
  \label{tbl:accuracy}
  \begin{tabular}{l|rrrr|r} \hline
    \multirow{2}{*}{Method} & \multicolumn{2}{c}{ModelNet40} &
      \multicolumn{2}{c|}{ScanObjectNN} & \multirow{2}{*}{\#Param} \\
    & mAcc(\%) & OA(\%) & mAcc(\%) & OA(\%) & \\ \hline
    PointNet~\cite{CharlesRQi17A} & 86.0 & 89.2 & 63.4 & 68.2 & 3.48M \\
    PointNet++~\cite{CharlesRQi17B} & 88.4 & 90.7 & 75.4 & 77.9 & 1.48M \\
    PointCNN~\cite{YangyanLi18} & 88.1 & 92.5 & 75.1 & 78.5 & -- \\
    PointConv*~\cite{WenxuanWu19} & -- & 92.5 & -- & -- & 18.6M \\
    KPConv~\cite{HuguesThomas19} & -- & 92.9 & -- & -- & 14.3M \\
    DGCNN~\cite{YueWang19} & 90.2 & 92.9 & 73.6 & 78.1 & 1.84M \\
    ExNODE~\cite{YangLi20} & -- & 89.3 & -- & -- & 0.52M \\
    PAConv~\cite{MutianXu21} & -- & 93.9 & -- & -- & 2.44M \\
    PCT~\cite{MengHaoGuo21} & -- & 93.2 & -- & -- & 2.88M \\
    PT*~\cite{HengshuangZhao21} & 90.6 & 93.7 & -- & -- & -- \\
    RepSurf~\cite{HaoxiRan22} & 91.4 & 94.4 & 81.3 & 84.3 & 1.48M \\
    PointNeXt-S~\cite{GuochengQian22} & 90.8 & 93.2 & 85.8 & 87.7 & 1.4M \\ \hline
    PointMLP & 91.0 & 93.5 & 83.5 & 85.3 & 13.24M \\
    PointMLP-Elite & 90.5 & 93.1 & 82.2 & 84.2 & 0.72M \\ \hline
    {\PointODENaive} & 89.4 & 92.9 & 82.5 & 84.3 & 7.66M \\
    {\PointODEElite} & 90.5 & 93.4 & 82.6 & 84.2 & 0.58M \\ \hline
  \end{tabular}
\end{table}

Table \ref{tbl:accuracy-ablation} presents the ablation results on ScanObjectNN to investigate the impact of the proposed architectural changes.
Compared to naively applying Neural ODE (Sec. \ref{sec:method-ode-block}), reordering the residual blocks (Sec. \ref{sec:method-reordering}, Fig. \ref{fig:pointmlp-reorder}) allows for 1.56x parameter reduction with an accuracy loss of 0.6\%.
The feature dimension reduction (Sec. \ref{sec:method-elite}) leads to 8.52x parameter savings without significantly harming the accuracy.
The point-wise normalization (Sec. \ref{sec:method-local-norm}) enables the pipelined feature extraction (Fig. \ref{fig:impl-stage}) and also improves the accuracy by 1.1--1.5\% by computing the normalizing factor independently for each point feature.

\begin{table}[htbp]
  \centering
  \caption{Ablation study results on ScanObjectNN.}
  \label{tbl:accuracy-ablation}
  \begin{tabular}{l|ccc|rr} \hline
    & \multirow{2}{*}{Reorder} & Point-wise & Reduced &
      \multicolumn{2}{c}{ScanObjectNN} \\
    & & Norm. & Dims. & mAcc(\%) & OA(\%) \\ \hline
    {\PointODENaive} & & & & 82.5 & 84.3 \\
    & $\checkmark$ & & & 81.9 & 83.8 \\
    {\PointODE} & $\checkmark$ & $\checkmark$ & & \textbf{83.4} & \textbf{84.9} \\
    & $\checkmark$ & & $\checkmark$ & 81.2 & 83.1 \\
    {\PointODEElite} & $\checkmark$ & $\checkmark$ & $\checkmark$ & 82.6 & 84.2 \\ \hline
  \end{tabular}
\end{table}

Table \ref{tbl:accuracy-zcu104} shows the accuracy of FPGA-based {\PointODEElite}.
While the FPGA implementation uses a fixed-point data type for point clouds and extracted features, it maintains nearly the same accuracy as the PyTorch counterpart on both datasets.
This suggests more aggressive quantization techniques could be applied to {\PointODEElite}, which is left as a future work.
Note the reduction of the mean accuracy by 0.5--0.6\% indicates the model makes slightly more incorrect predictions for categories with fewer training samples.

\begin{table}[htbp]
  \centering
  \caption{Classification accuracy on the ZCU104 board.}
  \label{tbl:accuracy-zcu104}
  \begin{tabular}{lc|rrrr} \hline
    \multirow{2}{*}{Method} & \multirow{2}{*}{FPGA} &
      \multicolumn{2}{c}{ModelNet40} & \multicolumn{2}{c}{ScanObjectNN} \\
    & & mAcc(\%) & OA(\%) & mAcc(\%) & OA(\%) \\ \hline
    {\PointODEElite} & & 90.5 & 93.4 & 82.6 & 84.2 \\
    {\PointODEElite} & $\checkmark$ & 90.0 & 93.7 & 82.0 & 84.3 \\ \hline
  \end{tabular}
\end{table}

\begin{figure}[htbp]
  \centering
  \includegraphics[keepaspectratio, width=0.5\linewidth]{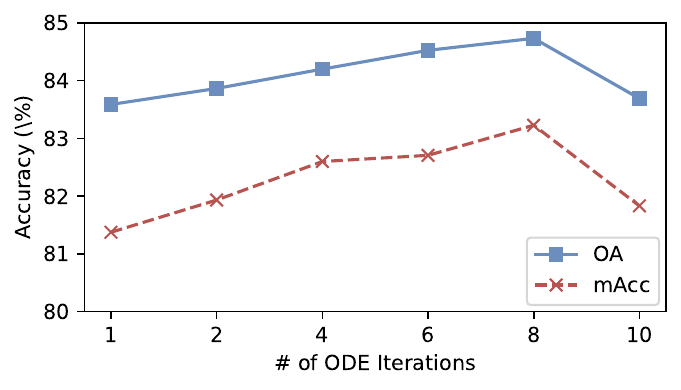}
  \caption{Number of ODE forward iterations $C$ and accuracy of {\PointODEElite} on the ScanObjectNN dataset.}
  \label{fig:acc-vs-iter}
\end{figure}

Fig. \ref{fig:acc-vs-iter} plots the accuracy of {\PointODEElite} with respect to the number of ODE iterations $C$ on ScanObjectNN.
By increasing $C$ from 1 to 8, the overall accuracy improves by 1.1\% (83.6\% to 84.7\%) at the cost of linearly increasing computational cost in {\ODEPBlock}s, which closes the gap between {\PointODEElite} and PointMLP.
Setting $C \ge 10$ results in a sudden accuracy drop, which may be attributed to cumulative numerical errors in the ODE solver.
Higher-order ODE solvers (e.g., Runge-Kutta) could be used instead to alleviate this problem.

\subsection{Inference Time} \label{sec:eval-inference-time}
\begin{figure}[htbp]
  \centering
  \includegraphics[keepaspectratio, width=0.6\linewidth]{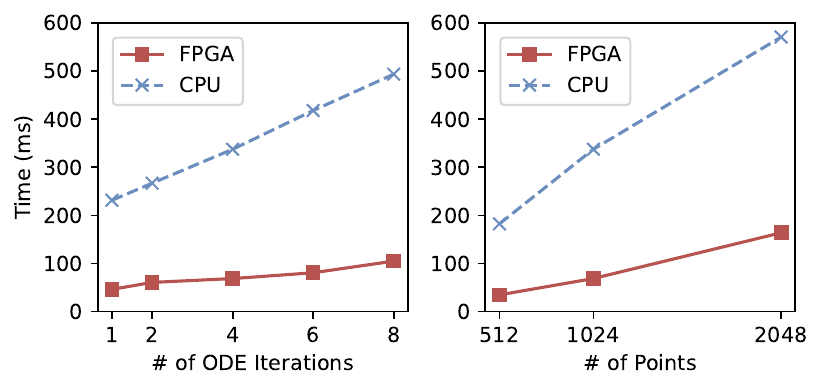}
  \caption{Time for feature extraction with respect to the number of ODE iterations $C$ (left) and number of points $N$ (right).}
  \label{fig:time-vs-iter-points}
\end{figure}

\begin{figure}[htbp]
  \centering
  \includegraphics[keepaspectratio, width=0.5\linewidth]{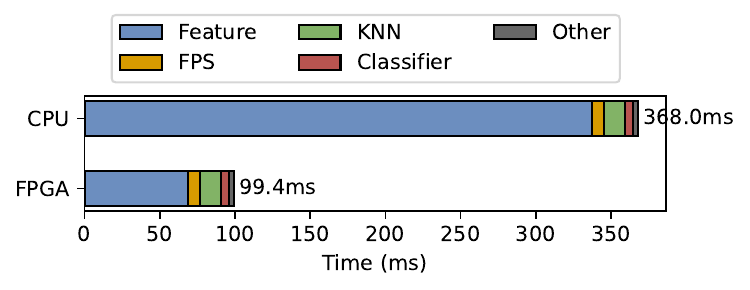}
  \caption{Execution time breakdown ($N = 1024, C = 4$).}
  \label{fig:time-bar}
\end{figure}

Fig. \ref{fig:time-vs-iter-points} shows the execution time of the feature extraction part in {\PointODEElite} with respect to $C$ and the number of points $N$.
Compared to the ARM Cortex-A53 CPU, the FPGA implementation achieves a speedup of 4.39--5.17x and 3.48--5.24x for $C = [1, 8]$ and $N = [512, 2048]$, respectively, thanks to the four-step feature extraction pipeline that processes four points in parallel (Fig. \ref{fig:impl-stage}).
The result indicates the linear computational complexity of {\ODEPBlock}s in terms of both $N$ and $C$.
Fig. \ref{fig:time-bar} compares the execution time breakdown of {\PointODEElite} ($N = 1024, C = 4$).
The feature extraction accounts for 91.7\% of the total inference time and dominates the performance.
The FPGA implementation accelerates feature extraction by 4.90x (from 337.5ms to 68.8ms), resulting in an overall speedup of 3.70x (from 368.0ms to 99.4ms).

\subsection{Power and Energy Consumption} \label{sec:eval-power}
The power consumption of {\PointODEElite} is calculated by subtracting that of the ZCU104 board in the idle state from that during inference.
The power usage is read out from an onboard INA226 sensor via the PMBus interface.
We repeat the inference 300 times and average the power measurements collected at 500ms intervals.
PyTorch- and FPGA-based {\PointODEElite} consume 0.43W and 0.45W of power (from 11.35W to 11.78W and 11.80W) during inference, respectively.
Considering the overall speedup of 3.70x (Sec. \ref{sec:eval-inference-time}), the FPGA implementation achieves 3.54x higher energy efficiency than the PyTorch counterpart.

\subsection{FPGA Resource Utilization} \label{sec:eval-resource}
Table \ref{tbl:resource} shows the FPGA resource utilization of {\PointODEElite}.
The design consumes 95\% and 69\% of the URAM and BRAM slices to store point clouds, extracted features, as well as parameters on-chip, eliminating most of the off-chip memory accesses.
The network compression techniques (e.g., quantization and pruning) would further reduce the BRAM and URAM utilization, which in turn allows to assign more memory blocks to each buffer and increase the number of read/write ports.
Considering that DSP blocks are underutilized, the design could further parallelize the computation in each module and save the inference time (e.g., each stage can process multiple sampled points at once).
More design optimizations should be explored in future work to fully utilize the computational capability of FPGAs.

\begin{table}[htbp]
  \centering
  \caption{FPGA resource utilization.}
  \label{tbl:resource}
  \begin{tabular}{l|rrrrr} \hline
    & BRAM & URAM & DSP & FF & LUT \\ \hline
    Total & 312 & 96 & 1728 & 460800 & 230400 \\
    Used & 215.5 & 92 & 731 & 80871 & 107028 \\
    Used (\%) & 69.1\% & 95.8\% & 42.3\% & 17.6\% & 46.5\% \\ \hline
  \end{tabular}
\end{table}


\section{Conclusion} \label{sec:conc}
In this paper, we propose {\PointODE} as an efficient DNN architecture for point cloud processing, along with its FPGA implementation.
{\PointODE} consists of a stack of residual MLP blocks to perform hierarchical feature extraction and aggregation.
The key idea behind {\PointODE} is to employ Neural ODE as a network compression technique.
We effectively reduce the network size by replacing a sequence of residual blocks with an ODE-based building block.
By adjusting the network architecture, we propose {\PointODEElite} as a lightweight version with only 0.58M trainable parameters, which is implemented on a Xilinx ZCU104 board.
We design a four-step feature extraction pipeline leveraging point-level parallelism and utilize on-chip memory to eliminate most of the off-chip data transfers.
While {\PointODEElite} does not employ sophisticated local feature extractors, its FPGA implementation achieves an accuracy of 93.7\% on the ModelNet40 dataset, which is even comparable to state-of-the-art networks.
In addition, it speeds up the inference by 3.7x and improves the energy efficiency by 3.5x compared to the PyTorch implementation.

\renewcommand{\baselinestretch}{1.0}
\bibliographystyle{unsrt}


\end{document}